\documentclass[letterpaper]{article} 
\usepackage{aaai23}  
\usepackage{times}  
\usepackage{helvet}  
\usepackage{courier}  
\usepackage[hyphens]{url}  
\usepackage{graphicx} 
\usepackage{natbib}  
\usepackage{caption} 
\usepackage{algorithm}
\usepackage{algorithmic}

\usepackage{amsmath}
\usepackage{amssymb}
\usepackage{bbm}
\usepackage{amsthm}

\DeclareMathOperator*{\argmin}{arg\,min}

\newtheorem{proposition}{Proposition}
\newtheorem{proposition_appx}{Proposition}

\usepackage{newfloat}
\usepackage{listings}
\DeclareCaptionStyle{ruled}{labelfont=normalfont,labelsep=colon,strut=off} 
\floatstyle{ruled}
\newfloat{listing}{tb}{lst}{}
\floatname{listing}{Listing}
\usepackage{booktabs}       
\usepackage{xcolor}         

\title{Losses over Labels: Weakly Supervised Learning via Direct Loss Construction}

\author {
    Dylan Sam\textsuperscript{\rm 1}, 
    J. Zico Kolter \textsuperscript{\rm 1} \textsuperscript{\rm 2}
}
\affiliations {
    \textsuperscript{\rm 1}   Machine Learning Department, Carnegie Mellon University\\
    \textsuperscript{\rm 2} Bosch Center for Artificial Intelligence \\
    \texttt{dylansam@andrew.cmu.edu}, \texttt{zkolter@cs.cmu.edu}
}

\begin{document}

\maketitle

\begin{abstract}    

Owing to the prohibitive costs of generating large amounts of labeled data, programmatic weak supervision is a growing paradigm within machine learning. In this setting, users design heuristics that provide noisy labels for subsets of the data. These weak labels are combined (typically via a graphical model) to form pseudolabels, which are then used to train a downstream model. In this work, we question a foundational premise of the typical weakly supervised learning pipeline: given that the heuristic provides all ``label" information, why do we need to generate pseudolabels at all? Instead, we propose to directly transform the heuristics themselves into corresponding loss functions that penalize differences between our model and the heuristic. By constructing losses directly from the heuristics, we can incorporate more information than is used in the standard weakly supervised pipeline, such as \emph{how} the heuristics make their decisions, which explicitly informs feature selection during training. We call our method Losses over Labels (LoL) as it creates losses directly from heuristics without going through the intermediate step of a label. We show that LoL improves upon existing weak supervision methods on several benchmark text and image classification tasks and further demonstrate that incorporating gradient information leads to better performance on almost every task.

\end{abstract}

\section{Introduction}

Recent advances in deep learning are enabled by the availability of large labeled datasets. However, expertly labeled data can be very costly to obtain, causing a bottleneck in many deep learning applications. Fortunately, in the absence of labeled data, we can leverage domain knowledge or auxiliary information for a given task. Many subfields of machine learning have tackled this problem, including semi-supervised learning, weakly supervised learning, and self-supervised learning.  Although the specific methods employed by these approaches are quite distinct, many of them operate under a common theme. These approaches often work by generating \textbf{pseudolabels} \citep{Lee2013PseudoLabelT, CascanteBonilla2020CurriculumLS, Pham2021MetaPL}, which can be plugged into a standard supervised learning pipeline to optimize a model on large amounts of unlabeled data. This paper focuses on their particular use in the paradigm of programmatic weak supervision \citep{zhang2022survey}, which we will refer to as weak supervision throughout this paper. 

In the setting of weak supervision, we assume the presence of \emph{weak labelers} that are commonly hand-engineered heuristics (we use the terms weak labeler and heuristic interchangeably, although weak labelers are more general and can have more varied structures). As an example, consider a sentiment classification task for restaurant reviews; a heuristic might associate the word ``delicious'' with a positive label, although this is clearly an imperfect label. Most recent advancements in weak supervision propose approaches that aggregate the outputs of multiple such heuristics to produce pseudolabels, typically through a graphical model \citep{snorkel}. These pseudolabels are used in combination with unlabeled data to optimize a downstream model, commonly referred to as an \textbf{end model}. This line of work has demonstrated its widespread applicability and efficiency \citep{drybell} to learn from noisy information.

\begin{figure*}[t]
    \centering
    \includegraphics[width=1.3\columnwidth]{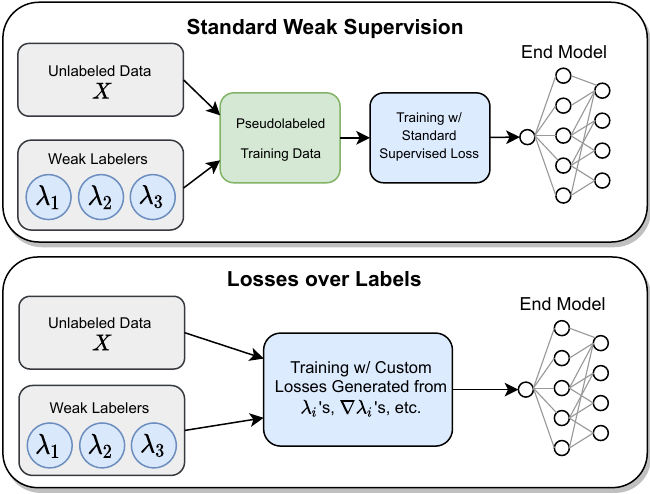}
    \caption{Depiction of our LoL method (bottom), compared to a standard weakly supervised pipeline (top). Our loss functions are generated by Equation \ref{grad_pen}.}
    \label{fig:lol_fig}
\end{figure*}

In this paper, we question this fundamental approach to weak supervision: specifically, why is there any need to produce pseudolabels at all? This standard approach distills the heuristics, which are rich sources of information, entirely into a single pseudolabel, to fit into a standard supervised learning pipeline. Instead, we propose to \emph{directly transform heuristics into loss functions} and train a network to minimize a combination of these loss functions (Figure \ref{fig:lol_fig}). We refer to our method as Losses over Labels (LoL), as we produce a combination of loss functions rather than intermediate pseudolabels. The simplest form of these losses is a smoothed variant of the number of times our model's prediction differs from the heuristic's prediction;
this is similar to a form of ``soft'' pseudolabels. Building on this, our losses incorporate additional information from the heuristics, such as \textit{which features are used in their decisions}. We add a penalty term to each loss, penalizing a model when its gradients do not match the heuristics' gradients on these important features. Incorporating this gradient information improves feature selection during optimization, which has not been previously considered in weak supervision and which indeed cannot be naturally performed with pseudolabels alone. Intuitively, this gradient information constrains our hypothesis space, removing lots of incorrect hypotheses that do not use correct features. Since our method uses the same heuristics as other weakly supervised algorithms, it does not require any additional information over existing methods; it just accesses the heuristic's underlying decision mechanism. 

We demonstrate that LoL performs favorably when compared to methods that create intermediate pseudolabels on five standard tasks from a weak supervision benchmark \citep{wrench}. 
Our results support that our approach to incorporate gradient information especially helps in the regime when we do not have access to abundant unlabeled data. As other methods have to implicitly learn important features from data, LoL directly incorporates some of this information from the heuristics and, thus, requires less unlabeled data.
Finally, we demonstrate that even when this gradient information is implicitly learnt (i.e., pretrained models on an auxiliary task), incorporating it during training still is beneficial to performance. As a whole, our results suggest that weakly supervised methods do not need to generate pseudolabels and can be further improved by directly translating heuristics into more informative loss functions to be optimized.  

\section{Related Work}

The problem of learning from weak or noisy sources of information has been studied in many different contexts. 

\textbf{Noisy Labels}\: Various approaches have tackled the problem of learning from a single source of noisy labels. Many of these techniques assume that these labels are corrupted with random noise \citep{noisy_labels, improve_lab_noise} and provide methods to address this in the optimization procedure or average over multiple splits of the dataset to denoise these corrupted labels. More recent techniques consider class-conditional noise or noise that is instance-dependent \citep{han2020survey} and provide techniques to correct such noise in the multiclass setting \citep{patrini2017making}. However, a fundamental distinction from this line of work is that weak supervision considers multiple sources of such noisy information.

\textbf{Crowdsourcing}\: Learning from multiple noisy sources has been classically studied in the field of crowdsourcing, where the goal is to combine human-annotated labels from multiple sources of varying quality. Many works in crowdsourcing combine these weak signals to recover the true labels. These approaches \citep{dawidskene, optimalindependence, gao2013minimax} make strong assumptions about the noisy annotators and derive optimal approaches to combine them under these assumptions. Prior work has extended this setting to allow abstentions \citep{aggregating_bin_ratings}. More recent crowdsourcing work has connected the field to modern machine learning applications by analyzing models trained on noisy crowdsourced labels \cite{learning_crowdsourcing}. All of these works focus on aggregating crowdsourced annotators into a pseudolabel to potentially optimize a downstream model, while our approach proposes generating loss functions.

\textbf{Ensemble Methods} Aggregating multiple functions has also been studied in the field of ensemble methods. In boosting \cite{schapire:ml90, adaboost, xgboost}, methods aggregate multiple weak learners into a strong classifier. Bagging \cite{breiman:ml96} aggregates multiple models to reduce variance from the individual (potentially overfitting) classifiers through bootstrapped samples. Ensemble methods have also been specifically studied in the semi-supervised setting \cite{balsubramani_scalable, balsubramani_optimally}, creating a combination of classifiers via a minimax game. While these are all similar to combining weak labelers, their focus is to produce a weighted combination of multiple weak learners rather than training a downstream model. 

\textbf{Multi-task Learning}\: Multi-task learning is a related field that focuses on optimizing multiple loss functions, which are derived from different tasks. Here, approaches train a single model on multiple learning tasks simultaneously \citep{multitask}. Many approaches have focused on formalizing a notion of task similarity \citep{exploiting_task_relatedness, task_sim_mtl} to guide learning. Recent applications of multi-task learning to neural networks use some form of parameter sharing \citep{multitask_feature, trace_reg_multi, sluice_nets, task_sim_mtl}. 
We remark that multi-task learning is different from our setting as our weak labelers are designed for the \textit{same task}, just with differing qualities and coverages. While a connection between weak supervision and multi-task learning has been made in previous work \citep{multi_task_weak_supervision}, this approach still combines information from multiple tasks into a single pseudolabel via a graphical model, rather than considering the heuristics individually.

\textbf{Weak Supervision}\: The paradigm of weak supervision assumes the presence of weak labelers of varying quality that label subsets of the data. A seminal work \citep{data_programming} first formulated combining these noisy sources of information via a graphical model, which produces pseudolabels used to train an end model.
Most recent advances in weakly supervised learning have trended towards producing higher quality pseudolabels, frequently through more complex graphical models \citep{multi_task_weak_supervision, fast_three}.
Additional works have developed variants of these graphical models to extend the weakly supervised setting to handle other objectives \citep{wiser, shin2022universalizing} and more general forms of weak supervision \citep{learning_partial, zhang2022creating}. Another line of weak supervision research has improved these methods by assuming additional information, i.e, a semi-supervised setting \cite{exemplars, dpssl}, an active learning setting \citep{biegel2021active}, or prior information on weak labeler accuracies \citep{gALL, AMCL, pgmv, Arachie2021ConstrainedLF, Arachie2022DataCF}. Finally, other works have developed more elaborate downstream training methods, primarily focused on deep learning models \citep{selftrain_wsl, endtoend, yu2021fine}. Overall, recent advances in weak supervision have focused on improving pseudolabels and potentially assuming access to more information. We remark that our method looks to improve learning from weak supervision in an orthogonal direction, by generating loss functions instead of pseudolabels.

\section{Preliminaries}

Our classification task has some domain $\mathcal{X}$ and a set of multiple discrete labels $\mathcal{Y}$ where $|\mathcal{Y}| = k$. We assume that data is given by some underlying distribution $\mathcal{D}$ over $\mathcal{X} \times \mathcal{Y}$. Our goal is to learn a classifier $h: \mathcal{X} \to [0, 1]^{k}$ that maps examples to a probability distribution over labels $\mathcal{Y}$, which minimizes standard 0-1 risk (on the argmax of the outputs of $h$). 

In the weakly supervised paradigm, we only observe training examples $X = \{x_1, ..., x_n \} \subset \mathcal{X}$. In the absence of labels for these examples, weak labelers are given as a set of $m$ functions $\lambda = \{\lambda_1, ..., \lambda_m \}$, and each $\lambda_i : \mathcal{X} \to \mathcal{Y} \cup \{ \emptyset \}$ where $\emptyset$ denotes an abstention. Abstentions allow more flexibility in the design of weak labelers, as users can create weak labelers that have high accuracy on specific regions of the data and abstain everywhere else. While weak labelers are permitted to abstain on data points, our classifier $h$ is not able to abstain and must always make a prediction for all $x$. 


\section{Losses over Labels}

Our main contribution is that, instead of producing pseudolabels, we propose to optimize a combination of loss functions that are derived directly from the weak labelers. This approach is arguably simpler than previous graphical models and still captures all ``label" information used to generate these pseudolabels. In fact, this allows us to retain more information from these multiple sources (such as which features inform their decisions) during the optimization procedure. Intuitively, this improves the training process by aiding (and potentially correcting) feature selection with gradient information from the heuristics. We remark that this information is ignored by existing weak supervision methods that aggregate heuristics to produce intermediate pseudolabels. 

\subsection{Generating Losses from Heuristics}

Formally, our approach looks to optimize an aggregation of loss functions that are derived from our heuristics. We can find some classifier $\hat{h}$
\begin{equation}\label{autoloss}
    \hat{h} = \argmin_h \:  \sum_{j=1}^n \Bigg( \frac{1}{m(x_j)}  \sum_{i=1}^m \ell_i(x_j, h) \Bigg),
\end{equation}
where $\ell_i$ corresponds to the loss function generated from weak labeler $\lambda_i$, and $m(x) = \sum_{i=1}^m \mathbbm{1}\{\lambda_i(x) \neq \emptyset\}$ represents the number of labelers that do not abstain on point $x$. Perhaps the simplest loss function that we can consider is the loss $\ell_i$ on data $x$ and classifier $h$ as \begin{equation}\label{simple_loss}
    \ell_i(x, h) =  \mathbbm{1}\{ \lambda_i(x) \neq \emptyset \} \cdot \ell(h(x), \lambda_i(x)),
\end{equation}
where $\ell$ can be any arbitrary loss function. This simple loss function can be easily generated given access to the heuristic function outputs. A combination of these losses represents the average loss $\ell$ over weak labelers that do not abstain. 
This is reasonable as our learnt model should conflict the fewest possible number of times with the weak labels. For the remainder of this paper, we consider $\ell$ to be the cross-entropy loss, which serves as a smoothed version of the number of disagreements between the trained classifier $h$ and weak labeler $\lambda_i$. We remark that this simple combination of losses is similar to optimizing over pseudolabels with a ``soft" majority vote function. In fact, under the log loss and the square loss, we recover the same objective up to a constant factor. 

\begin{proposition}\label{log_loss}
 An aggregation of log losses is equivalent to using a soft version of a majority vote as a pseudolabel.
\end{proposition}
\begin{proposition}\label{sq_loss}
 An aggregation of squared losses is equivalent to using a soft version of a majority vote as a pseudolabel, up to an additive constant (with respect to our model).
\end{proposition}

The derivations of these propositions are shown in Appendix \ref{minimizers}. While this combination of naive losses recovers the same objective as using pseudolabels, this overall framework paves the road for us to create more informative and powerful loss functions from heuristics by incorporating their gradient information. 

\subsection{Incorporating Additional Heuristic Information via Input Gradients}

We now present our approach to incorporate gradient information from the heuristics, in order to produce more complex losses. In the typical weak supervision pipeline, the weak labeler aggregation model only looks at the \textit{outputs of the weak labelers}. As the weak labelers are hand-engineered by domain experts, we frequently have access to the underlying mechanism of the heuristics. Therefore, we have information about which features are used to make their decisions. 

The most common form of weak labelers (in a text classification setting) is that of a rule that checks for the presence of a set of words in a given example. For a bag of words data representation, we can determine exactly which dimensions of the data $x$ our weak labelers use, by finding the indices of the words in the vocabulary. This is a rich source of information, which is thrown away when generating pseudolabels. Therefore, we propose to leverage these additional properties of the heuristics through their input gradients during optimization. 
We remark that this form of weak labelers is a binary-valued function and is, thus, not differentiable. Therefore, we can incorporate their ``gradient information" in a principled manner by creating a smoothed approximation of these heuristics via random inputs to the heuristics. This is similar to an idea used in adversarial robustness \citep{random_smoothing}. 

Formally, let $\mathcal{X} = \{0, 1 \}^n$ and $\mathcal{Y} = \{0, 1\}$, although this easily generalizes to a multiclass setting. Then, we can consider a heuristic function that looks for the presence of a word at index $j$, or $\lambda_i(x) = \mathbbm{1}_{\{x_j = 1\}}$ and provides a positive label if it is present.  We can define the smoothed approximation of this heuristic $\tilde{\lambda}_i: [0, 1]^n \to [0, 1]^{k+1}$ as:
\begin{equation*}
    \tilde{\lambda_i}(\phi) = E_{x \sim Ber(\phi)} [ \lambda_i(x)],
\end{equation*}
where $\phi = (\phi_1, ..., \phi_n ) \in [0, 1]^n$. This is now a (continuous-valued) smoothed approximation of the heuristic by considering its expectation over random independent Bernoulli inputs to the function. Then $x_j = 1$ with probability $\phi_j$, which implies that 
\begin{equation*}
    \tilde{\lambda}_i(\phi_j) = \begin{pmatrix} 
    0 \\
    \phi \\
    1 - \phi
\end{pmatrix},
\end{equation*} 
where the first index corresponds to class 0, the second index corresponds to class 1, and the third index corresponds to the abstain vote $\emptyset$. We can now analytically compute a gradient of the smoothed approximation of our weak labeler $\tilde{\lambda}_i$ with respect to the Bernoulli distribution parameter $\phi$. The gradient of the smoothed heuristic $i$ at index $j$ is given by 
\begin{equation*}
    \nabla_{\phi_j} \: \tilde{\lambda}_i(\phi_j) = \begin{pmatrix} 0 \\
    1 \\
    -1  \end{pmatrix}.
\end{equation*}
However, the binary nature of these heuristics only allows for gradients of the smoothed approximations to take values of $-1$ or $1$. As our learnt classifier takes real-valued inputs and produces real-valued outputs, we penalize our model for having an input gradient less than some hyperparameter $c$ times $\nabla_{\phi} \: \tilde{\lambda_i}$, which serves as a threshold for our model's desired gradient. In essence, this penalizes models that do not use the same features that are used by the heuristic. 

While we can analytically compute the randomized approximation of weak labelers, we can only approximate the gradient of our classifier. We can similarly define a smoothed approximation of our classifier $h$ on random Bernoulli inputs as
\begin{equation*}
    \tilde{h}(\phi) = E_{x \sim Ber(\phi)}[h(x)],
\end{equation*}
and we can estimate the input gradient empirically as $\nabla_{\phi} \: \tilde{h}(\phi)= \frac{1}{t}\sum_{i=1}^t \nabla_{z_i} h(z_i)$ where $z = \{z_1, ..., z_t\}$ is sampled iid from $Ber(\phi)$. Then, we can construct a gradient loss function $\ell_i^*$ induced by a weak labeler $\lambda_i$ as
\begin{equation}\label{grad_pen}
\begin{split}
    \ell^*_i(x, h) = \:   \mathbbm{1} \{\lambda _i(x) \neq \emptyset \} \cdot \Big( \ell(h(x), \lambda_i(x)) + \\ \alpha \cdot || \Big[ c  \cdot \nabla_{\phi} \tilde{\lambda}_i(\phi)_\mathcal{Y} - \nabla_{\phi}  \tilde{h}(\phi) \Big]^+ ||_{2}^2 \Big),    
\end{split}
\end{equation}

where $\tilde{\lambda}_i(\phi)_{\mathcal{Y}}$ denotes the gradient of the smoothed heuristic only along \textit{dimensions that correspond to non-abstaining votes} and where $[ x ]^+ = \max(x, 0)$, $\alpha > 0$, and $c > 0$. 
We note that can only match the gradients of the heuristic on the non-abstaining dimensions 
as our model $h$ cannot abstain.

At a high level, this loss function incorporates a squared penalty for the gradient of our model being less than $c$ times the gradient of the heuristic (along non-abstained dimensions). $\alpha$ serves as a hyperparameter that determines the weighting or importance of the gradient matching term, similar to a weighting parameter for regularization. We provide ablations in Appendix \ref{grad_ablation} to compare with other choices of this gradient penalty (e.g., linear or exponential). 

Finally, we can compute the empirical risk minimizer over the average of these gradient loss functions to produce a classifier $\hat{h}_g$:
\begin{align}\label{gradloss}
    \hat{h}_g = \argmin_h \sum_{j=1}^n \Bigg( \frac{1}{m(x_j)}\sum_{i=1}^m  \ell_i^*(x_j, h) \Bigg).
\end{align}
We remark that matching the input gradients of our learnt model to that of the heuristics only requires unlabeled data to compute. Therefore, our approach does not require any additional information, such as labeled data, to incorporate information about how the heuristics make their decisions. While this paper focuses on this prevalent type of heuristic, it generalizes to any weak labeler that has a gradient that we can compute. We demonstrate this applicability to other forms or weak labelers and types of data, such as pretrained models on image classification tasks, in Section \ref{image_experiments}.

\renewcommand{\arraystretch}{1.2}
\begin{table*}[t]
\centering
\caption{Results on 5 datasets from the WRENCH benchmark, averaged over 5 random seeds. Accuracies are reported as mean $\pm$ standard deviation. We note the best-performing method in red and the second best-performing method in blue. We omit results from T-Mean and T-Median on multiclass tasks (agnews, chemprot) as they are binary classification methods.}
\setlength{\tabcolsep}{5pt}
\resizebox{1.7\columnwidth}{!}{
\begin{tabular}{l | c c c c c | c }
\toprule
 & MV & Snorkel & T-Mean & T-Median & E2E & \textbf{LoL} \\ \midrule
agnews    & 82.3 $\pm$ 1.1 & \textbf{\textcolor{blue}{82.7 $\pm$ 0.9}} & -- & -- & 74.8 $\pm$ 0.1 & \textbf{\textcolor{red}{83.4 $\pm$ 0.1}} \\
chemprot  & \textbf{\textcolor{blue}{51.4 $\pm$ 0.4}} & 51.0 $\pm$ 0.4 & -- & -- & 50.1 $\pm$ 1.2 & \textbf{\textcolor{red}{52.9 $\pm$ 0.3}} \\
IMDB      & 80.8 $\pm$ 0.2 & \textbf{\textcolor{red}{82.1 $ \pm $ 0.5}} & 79.3 $ \pm $ 1.1 & 80.4 $\pm$ 0.8 & 75.2 $\pm $ 0.6 & \textbf{\textcolor{blue}{81.8 $\pm$ 0.3}} \\
Yelp      & 77.1 $\pm$ 1.1 & 77.7 $ \pm $ 0.5 & 79.3 $ \pm $ 1.1 & \textbf{\textcolor{red}{81.0 $\pm$ 1.0}} & \textbf{\textcolor{blue}{79.7 $\pm$ 0.2}} & 75.9 $\pm$ 0.7\\
YouTube   & \textbf{\textcolor{blue}{92.9 $\pm$ 0.9}} & 91.0 $\pm$ 0.3 & 90.0 $\pm$ 0.7 & 90.6 $\pm$ 0.5 & 91.7 $\pm$ 0.5 & \textbf{\textcolor{red}{94.2 $\pm$ 0.7}} \\ \bottomrule
\end{tabular}
}
\label{tab:results}
\end{table*}

\renewcommand{\arraystretch}{1.2}
\begin{table*}[t]
\centering
\caption{Results when training on only \textit{100 unlabeled data}, averaged over 5 random seeds. Accuracies are reported as mean $\pm$ standard deviation. Again, we note the best-performing method in red and the second best-performing method in blue.}
\setlength{\tabcolsep}{5pt}
\resizebox{1.7\columnwidth}{!}{
\begin{tabular}{l | c c c c c | c }
\toprule
 & MV & Snorkel & T-Mean & T-Median & E2E & \textbf{LoL} \\ \midrule
agnews    & \textbf{\textcolor{blue}{52.9 $\pm$ 1.8}} & 54.7 $\pm$ 1.8 & -- & -- & 46.7 $\pm$ 2.5  & \textbf{\textcolor{red}{63.2 $\pm$ 0.7}} \\
chemprot  & \textbf{\textcolor{blue}{38.2 $\pm$ 0.7}} & 36.5 $\pm$ 1.9 & -- & -- & 37.1 $\pm$ 1.1 & \textbf{\textcolor{red}{39.8 $\pm$ 1.2}} \\
IMDB      & \textbf{\textcolor{blue}{67.5 $\pm$ 0.6}} & 66.0 $ \pm $ 1.0 & 58.3 $ \pm $ 2.5 & 52.4 $\pm$ 0.7 & 61.7 $\pm$ 1.0 & \textbf{\textcolor{red}{69.6 $\pm$ 0.4}}\\
Yelp      & 67.8 $\pm$ 1.5 & 67.7 $\pm$ 2.2 &  \textbf{\textcolor{blue}{70.0 $ \pm $ 1.7}} & 65.0 $\pm$ 2.6 &  65.9 $\pm $ 1.4 & \textbf{\textcolor{red}{71.2 $\pm$ 2.4}} \\
YouTube   & \textbf{\textcolor{blue}{90.9 $\pm$ 1.01}} & 87.5 $\pm$ 2.4 & 88.0 $\pm$ 1.3 & 79.8 $\pm$ 4.5 & 90.2 $\pm$ 0.7 & \textbf{\textcolor{red}{92.0 $\pm$ 0.9}} \\ \bottomrule
\end{tabular}
}
\label{tab:results_ul}
\end{table*}

\section{Experiments}\label{Experiments}

In our experiments, we compare LoL to existing weakly supervised algorithms on 5 text classification datasets from WRENCH \citep{wrench}, which provides the weak supervision sources. For all of our text classification tasks, we use a bag of words data representation, so we can easily compute the gradients of a smoothed version of the heuristics. Most weak labelers in WRENCH check for the presence of a word or a set of words in a given sentence. We still include weak labelers that are of a different form (e.g., a pre-trained sentiment classifier) for all methods. However, we are unable to analytically compute the gradients for these other weak labelers, so we remove those terms from the LoL loss and use the simple loss given in Equation \ref{simple_loss}. In Section \ref{image_experiments}, we extend our setting to consider 3 image classification tasks from the Animals with Attributes 2 dataset \citep{xian:pami18} with weak labelers that are trained models. Code for our experiments can be found here\footnote{https://github.com/dsam99/LoL}.

\subsection{Baselines}

We compare our method against weakly supervised baselines that do not require any labeled data or estimates of weak labeler error rates. Our baselines include:

\textbf{Majority Vote (MV)}: We create pseudolabels with a vanilla majority vote where ties are broken randomly. These pseudolabels are then used to train an end model.

\textbf{Snorkel MeTaL (Snorkel)}: We generate soft pseudolabels via the graphical model approach in Snorkel MeTaL \citep{multi_task_weak_supervision}, which are used to train an end model.

\textbf{Triplet Methods (T-Mean and T-Median)}: We generate soft pseudolabels via the triplet method described in FlyingSquid \citep{fast_three}. Mean denotes the original approach, and median denotes the extension later for estimating median parameters \citep{Chen2021ComparingTV}. Both of these produce pseudolabels that are used to train an end model. We remark that these models are defined for binary classification tasks, so we cannot report their scores on multiclass tasks.

\textbf{End to End (E2E)}: We compare against the end-to-end approach that jointly learns both the aggregation of weak labelers and the end model \citep{endtoend}. We remark that this approach uses an additional neural network as an encoder to generate pseudolabels. Therefore, it uses a more complex model as compared to LoL and other baselines.

\subsection{Results}

We provide the results of our methods and the baselines in Table \ref{tab:results}.
LoL achieves the best performance on 3 of the 5 datasets. In addition, our method is the second best-performing method on another dataset. While T-Median and E2E are the best-performing methods on the remaining dataset, we note that LoL outperforms both of these methods on all other tasks. Even when compared to the much more complex architecture of E2E, LoL still performs favorably in most cases. 

These experiments illustrate that LoL achieves frequently better performance than existing weakly supervised approaches, without the need for intermediate pseudolabels. Furthermore, this demonstrates that using information about how the heuristics make their decisions can benefit weakly supervised methods, improving overall performance.

\begin{figure*}[t]
    \centering
    \includegraphics[width=0.6\columnwidth]{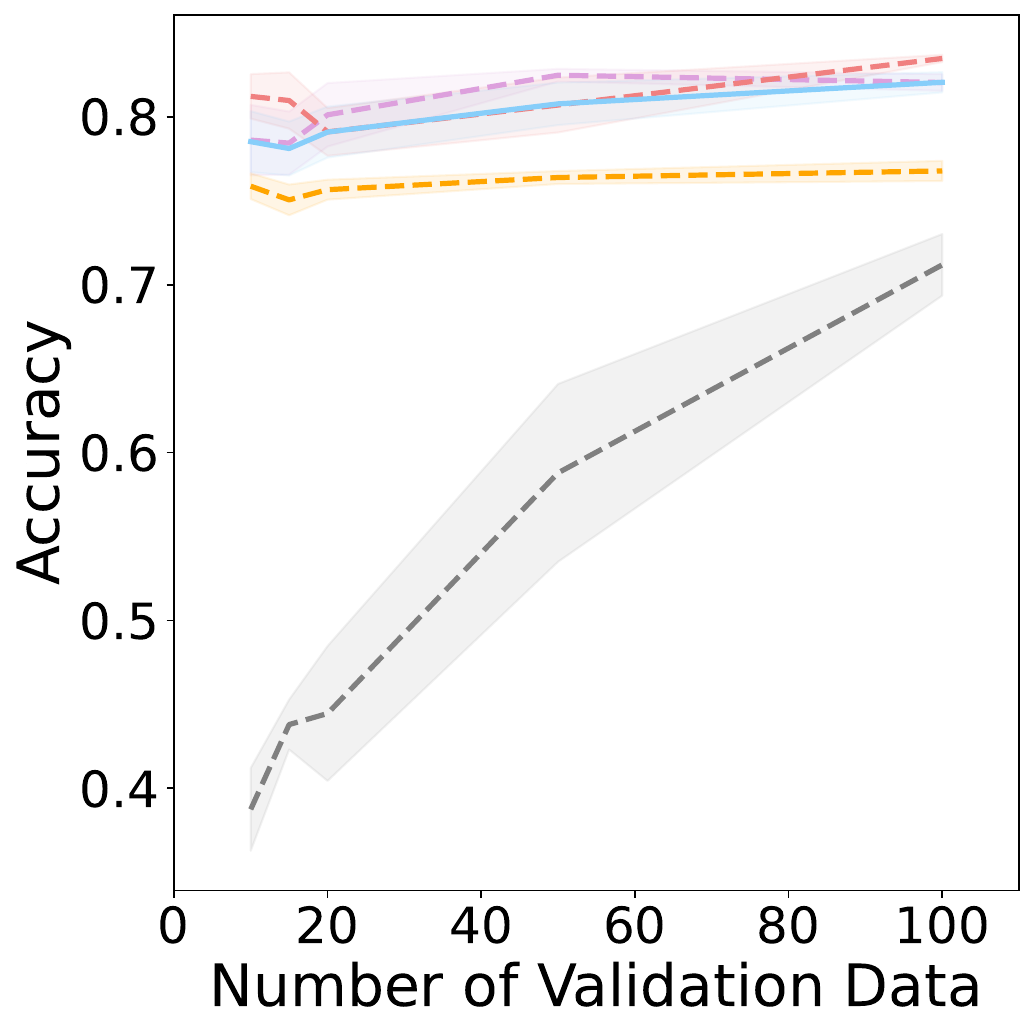}
    \includegraphics[width=0.6\columnwidth]{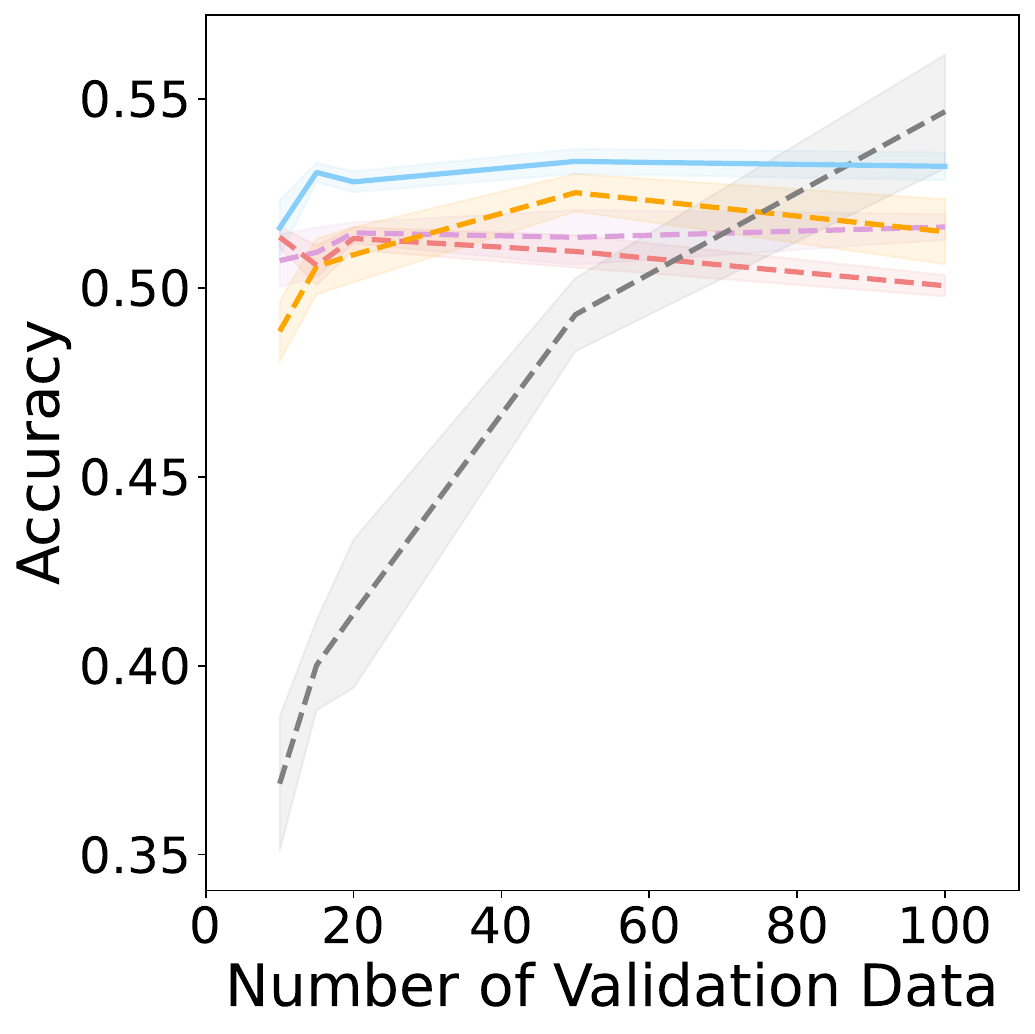} 
    \includegraphics[width=0.6\columnwidth]{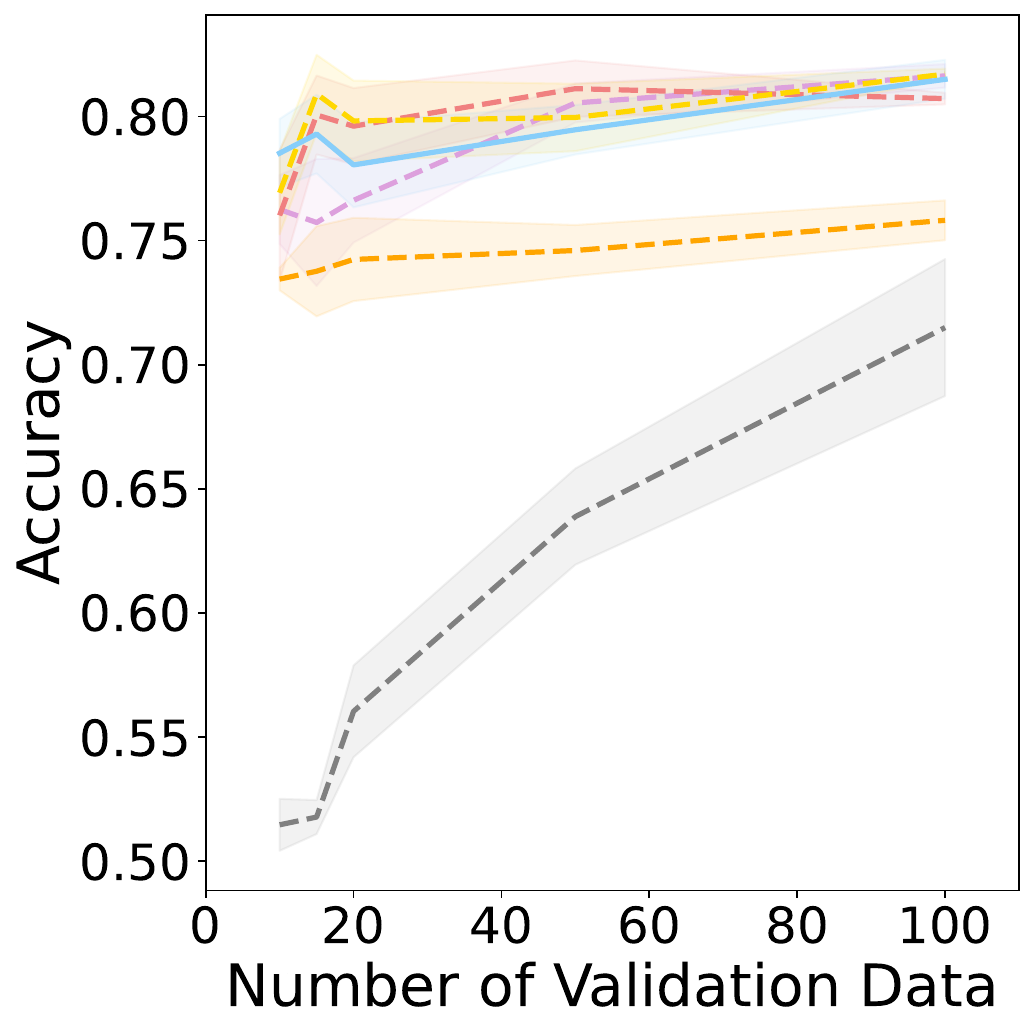}
    \includegraphics[width=0.6\columnwidth]{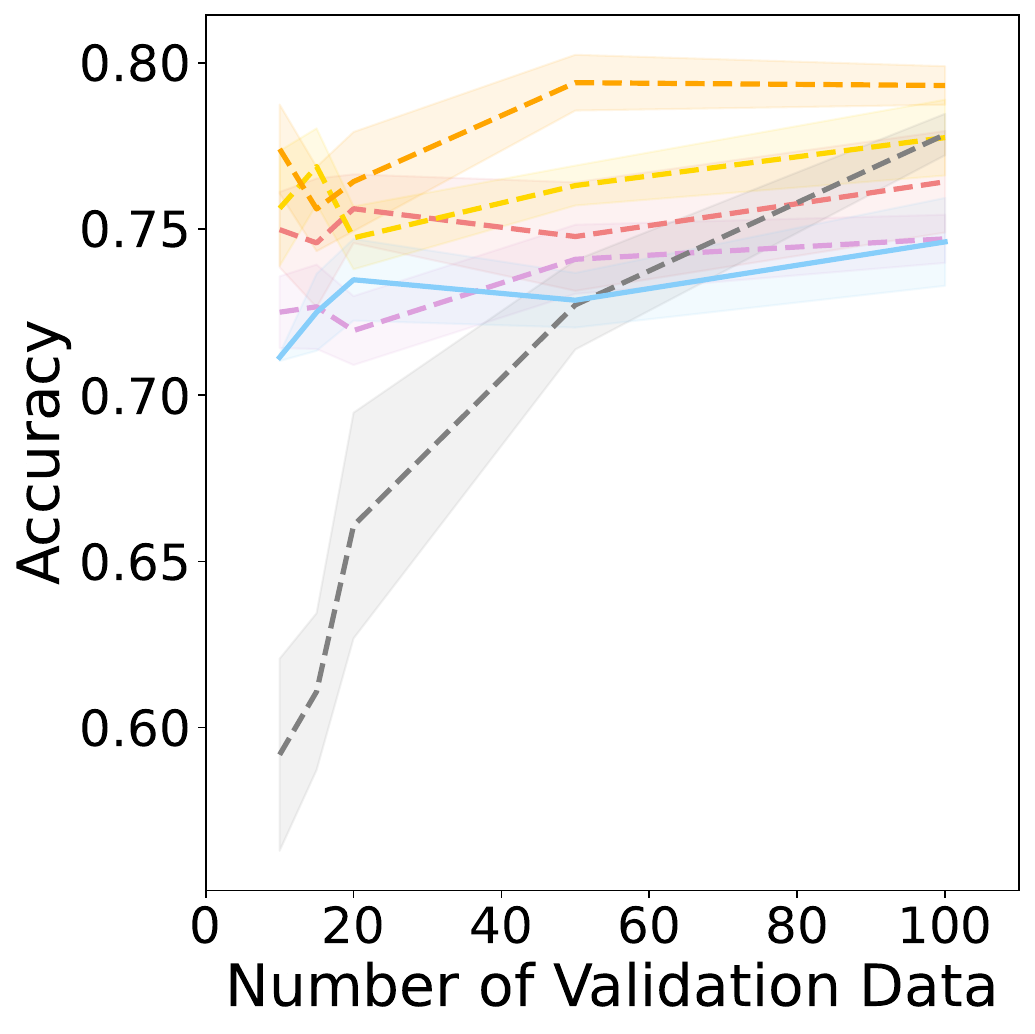}
    \includegraphics[width=0.6\columnwidth]{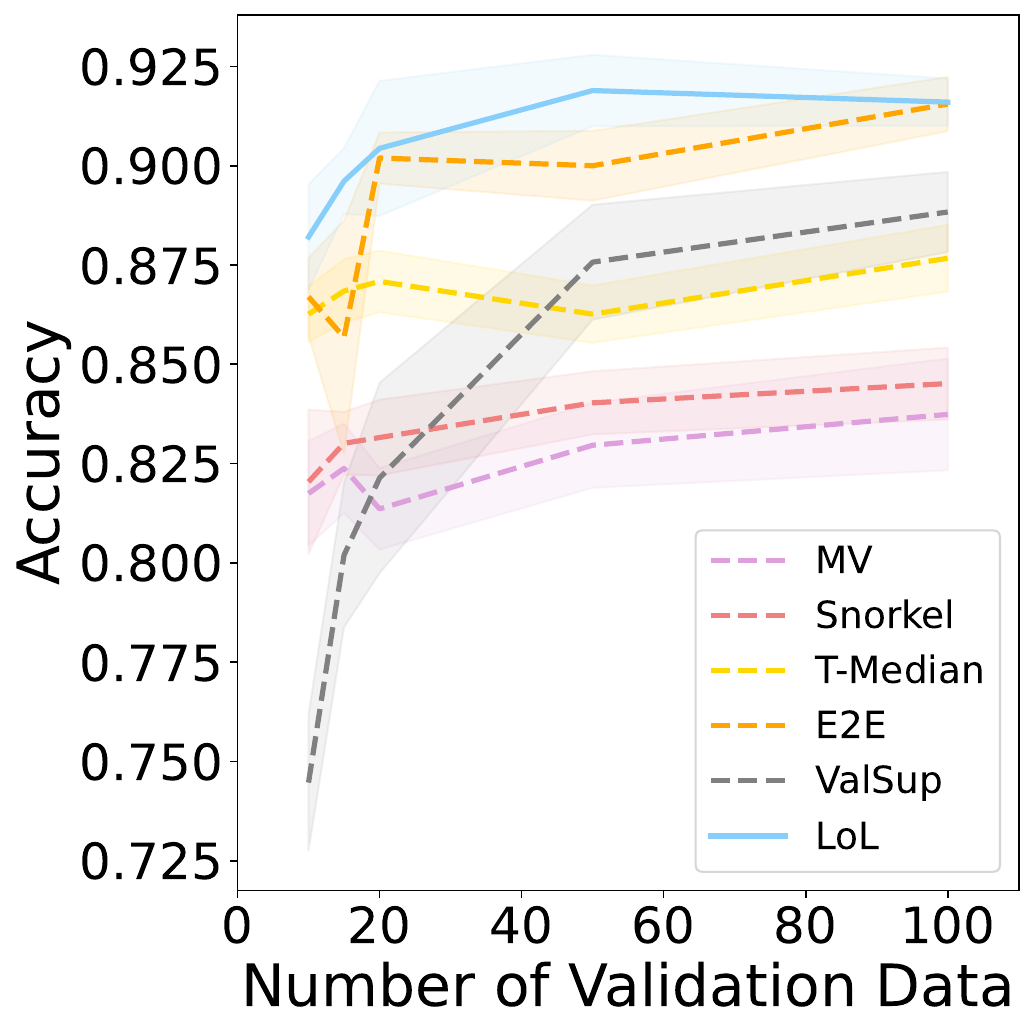}
    \caption{Results (left to right, top to bottom: agnews, chemprot, IMDB, Yelp, Youtube) as we vary validation set size. Baselines are given by \textit{dashed lines}, and LoL is given by a \textbf{solid line}. We compute error bars (standard deviation) over 5 seeds.}
    \label{fig:vary_lab}
\end{figure*}

\subsection{Performance with Limited Unlabeled Data}

While weakly supervised learning usually assumes abundant unlabeled data, we demonstrate the ability of LoL on settings with limited unlabeled data. We remark that LoL outperforms other methods, especially when unlabeled data is also limited in quantity. This is due to the additional information from how the heuristics make their decisions; this additional gradient information helps with feature selection during the training process. On the other hand, other methods must learn this information from unlabeled data, failing when unlabeled data is limited. 

We consider the same experimental setting as above in Table \ref{tab:results_ul}; however, we now only assume the presence of a few ($N = 100$) unlabeled training data from each task. We present these additional empirical results in Table \ref{tab:results_ul}. In this setting, LoL is now the best-performing method across all datasets. 
This is a notable result as LoL was previously outperformed on both IMDB and Yelp with large unlabeled data. In fact, we remark LoL with only 100 unlabeled data performs better than almost all other methods, using the full unlabeled training dataset (excluding MV).

The best performing baseline in most cases is MV, as other baselines (Snorkel, T, T-Median) likely produce incorrect accuracy estimates with limited unlabeled data. In addition, E2E is a more complex model, with an additional neural network as an encoder; thus, it struggles with learning from a small amount of data. These experiments demonstrate that our method especially improves performance specifically in the case when unlabeled data is limited.

\subsection{Impact of Using Gradient Information} \label{lol_vs_lol-simple}

We provide an ablation study in Table \ref{tab:simple_results} to directly compare using gradient information to the combination of naive losses (i.e, a soft majority vote pseudolabel), which we refer to as \textbf{LoL-simple}. This comparison looks to demonstrate the impacts of incorporating gradient information, which is the only difference between these two approaches.

\renewcommand{\arraystretch}{1.2}
\begin{table}[h]
\centering
\caption{Comparison of LoL and LoL-simple, averaged over 5 random seeds. Accuracies are reported as mean $\pm$ standard deviation. We bold the best performing method.}
\setlength{\tabcolsep}{5pt}
\resizebox{0.7\columnwidth}{!}{
\begin{tabular}{l | c c}
\toprule
 & \textbf{LoL} & \textbf{LoL-simple} \\ \midrule
agnews    &  \textbf{83.4 $\pm$ 0.1} & 82.3 $\pm$ 1.2 \\
chemprot  &  52.9 $\pm$ 0.3 & \textbf{53.2 $\pm$ 0.2} \\
IMDB      &  \textbf{81.8 $\pm$ 0.3} & 80.8 $\pm$ 0.8 \\
Yelp      &  \textbf{75.9 $\pm$ 0.7} & 74.8 $\pm$ 0.4 \\
YouTube   &  \textbf{94.2 $\pm$ 0.7} & 93.7 $\pm$ 0.8 \\ \bottomrule
\end{tabular}
}
\label{tab:simple_results}
\end{table}

\renewcommand{\arraystretch}{1.2}
\begin{table*}[t]
\centering
\caption{Results on binary classification tasks between test classes of the Animals with Attributes 2 dataset, averaged over 5 random seeds. We selected the 3 tasks with the fewest weak labelers (defined by differences in class attribute annotations). Accuracies are reported as mean $\pm$ standard deviation. $k = 20$ for LoL. The best-performing method is bolded.}
\setlength{\tabcolsep}{5pt}
\resizebox{2.02\columnwidth}{!}{
\begin{tabular}{l | c c c c | c | c c}
\toprule
 & MV & Snorkel & T-Mean & E2E & LoL-simple & \textbf{LoL} & \textbf{LoL-sw} \\ \midrule
seal v. whale & 90.2 $\pm$ 1.5 & \textbf{97.1 $\pm$ 0.6} & 93.1 $\pm$ 0.8 & 96.6 $\pm$ 0.6 & 91.8 $\pm$ 1.5 & 93.6 $\pm$ 1.3 & 95.5 $\pm$ 0.9 \\ 
raccoon v. rat & 75.3 $\pm$ 5.7 & 94.9 $\pm$ 0.9 & 85.1 $\pm$ 3.0 & 91.8 $\pm$ 2.1 & 69.9 $\pm$ 5.6 & 77.9 $\pm$ 4.1 & \textbf{96.4 $\pm$ 0.5} \\
whale v. hippo & 98.4 $\pm$ 0.4 & \textbf{99.1 $\pm$ 0.4} & 98.1 $\pm$ 0.8 & 98.4 $\pm$ 0.4 & 97.3 $\pm$ 0.5 & 98.3 $\pm$ 0.3 & 98.1 $\pm$ 0.4 \\ \bottomrule
\end{tabular}
}
\label{tab:results_awa2}
\end{table*}

We observe that in almost every dataset (except chemprot), incorporating gradient information improves upon the simple combination of labelers (that is equivalent to a soft majority vote). Only on the chemprot dataset, LoL-simple performs slightly better to LoL. This supports that gradient information is beneficial to the standard weakly supervised learning pipeline.

\subsection{Sensitivity to Amount of Validation Data}

Finally, we provide experiments that demonstrate the performance of LoL as we vary the amount of validation data used for model selection. We note that LoL has two additional hyperparameters $\alpha, c$; thus, it potentially requires more validation data. The WRENCH benchmark uses 10\% of the entire data as validation data, which is frequently sufficient labeled data to even train a supervised model. Therefore, we provide experiments (Figure \ref{fig:vary_lab}) that evaluate the sensitivities of our methods to the amount of validation data. 

For each task, we split the dataset into 80\% train and validation data and 20\% test data. Then we further split training and validation data into $N$ examples \textit{per class} of labeled validation data. We report results for validation set sizes of $N \in \{10, 15, 20, 50, 100\}$. We additionally compare all methods against a supervised learning approach on the labeled validation data (\textbf{ValSup}). This baseline helps us assess the particular regime of our experiment; we primarily care about when traditional supervised learning fails.
We note that the ValSup supervised learning baseline performs model selection on its own training set. In these experiments, we train methods for 10 epochs for early stopping given few validation data. 

Our results demonstrate that our model still performs comparably to or better than other methods, with access to little validation data. In fact, on the YouTube dataset (bottom-right) we outperform all methods, including the supervised baseline.

\subsection{Extension to Pretrained Labelers on Images} \label{image_experiments}
We extend our setting to consider image classification tasks with weak labelers that are trained models. We follow the standard procedure outlined in \cite{pgmv, AMCL} to create weak labelers from the Animals with Attributes 2 dataset \citep{xian:pami18}. These models learn to detect coarse-grained attributes from images (e.g., color, presence of water, etc.), which we use to produce noisy labels for new classes at test time. We follow the same experimental procedure as before, now using an underlying representation given by the pretrained ResNet18. 

To generate our gradient constraints, we can directly compute the input gradients of the weak labelers, as they are functions defined on continuous input data. However, as these models are learnt, we do not have any explicit control over their gradients. Therefore, we only constrain our model to match the top-$k$ gradient features of the weak labeler as they may be potentially noisy. Intuitively, we again benefit from additional feature selection as we add constraints that our models should explicitly use the most informative features for these auxiliary tasks, while not requiring the model to match less informative features.

We compare LoL against baselines on these image classification tasks in Table \ref{tab:results_awa2}.
We additionally provide a Snorkel-weighted variant of LoL, called \textbf{LoL-sw}. More discussion on different potential weightings of our losses is given in Appendix \ref{weightings}. We observe that our method is outperformed by Snorkel on two tasks but is the best performing method on one task. As before, we also observe that incorporating constraints based on gradient information does lead to benefits over a simple combination of soft weights, as LoL improves over LoL-simple on all tasks. This demonstrates that incorporating gradient information from weak labelers, \textit{even that which is implicitly learnt on auxiliary tasks}, can yield further improvements in the field of weak supervision.

\section{Conclusion}

In summary, we present a new and effective approach in the weakly supervised paradigm to optimize a model via a combination of loss functions that are generated directly from the weak labelers. Under this framework, these losses can leverage information about how the heuristics make decisions in our optimization, which has not been done previously by weakly supervised methods. We demonstrate that, on many text and image classification tasks, LoL achieves better performance than weakly supervised baselines and that incorporating (even learnt) gradient information almost always leads to better performance.

Our work has the potential to start a new direction of research in weakly supervised learning. While recent work in weak supervision has focused on improving the process of creating pseudolabels, our approach proposes a fundamentally different pipeline. When creating pseudolabels, we admittedly gain the benefit of using a plethora of standard supervised learning methods and models. However,  we lose the ability to impose additional control over the model optimization process. Incorporating more information from the heuristics directly in our method gives experts more control over the training procedure of the model. In fact, this result has larger implications on the design of the weak labelers themselves; experts now have a greater influence on the weakly supervised pipeline, as they can induce constraints on the gradients of the learnt model via designing heuristics.

\clearpage

\section*{Acknowledgements}

Dylan Sam was supported by funding from the Bosch Center for Artificial Intelligence and by the ARCS Foundation.

\bibliography{aaai23.bib}


\clearpage

\appendix

\section{Aggregation of Simple Losses} \label{minimizers}

We can consider the same setup as in the Preliminaries section, again focusing on the binary classification setting for simplicity ($\mathcal{Y} = \{ 0, 1\}$). We denote $a_p = a_p(x) = \sum_{i=1}^m \mathbbm{1}\{\lambda_i(x) = 1\}$ and $a_n = a_n(x) = \sum_{i=1}^m \mathbbm{1}\{\lambda_i(x) = 0\}$, which represent the number of positive (class 1) and negative (class 0) votes for a given data point $x$. We now analyze some concrete cases of an aggregation of simple loss functions (not including the gradient information used in our LoL method). We consider the log loss and the squared loss, and restate the propositions from the main text. 

\begin{proposition_appx}
A simple aggregation of log losses is equivalent to using a soft version of the majority vote function as a pseudolabel.
\end{proposition_appx}
\begin{proof}
We have that LoL under the log loss is given by
\begin{align*}
    \ell_{\text{LoL}}(x, h) &  = \frac{1}{m(x)} \sum_{i=1}^m \mathbbm{1}\{ \lambda_i(x) \neq \emptyset \} \cdot \Big( \lambda_i(x) \log h(x) + \\
     & \quad \quad \quad (1 - \lambda_i(x)) \log (1 - h(x)) \Big) \\
     & = \frac{1}{a_p + a_n} \Big( \sum_{i=1}^{a_p} \log(h(x)) + \sum_{j=1}^{a_n} \log (1 - h(x)) \Big) \\
     & = \frac{a_p}{a_p + a_n} \log h(x) + \frac{a_n}{a_p + a_n} \log (1 - h(x)) \\
     & = \ell_{\text{log loss}} \Big(h(x), \frac{a_p}{a_p + a_n} \Big),
\end{align*}

which is exactly the objective of the soft majority vote. 
\end{proof}

This shows that our simple aggregation of naive losses has an equivalent form to optimizing pseudolabels from a soft majority vote function. Next, we will consider the square loss.

\begin{proposition_appx}
 A simple aggregation of squared losses is equivalent to using a soft version of the majority vote function as a pseudolabel, up to an additive factor $\big( \frac{a_p}{a_p + a_n} - \big( \frac{a_p}{a_p + a_n} \big)^2\big)$.
\end{proposition_appx}

\begin{proof}

 We have that LoL under the log loss is given by
 \small
\begin{align*}
    \ell_{\text{LoL}}(x, h) &  = \frac{1}{m(x)} \sum_{i=1}^m \mathbbm{1}\{ \lambda_i(x) \neq \emptyset \} \Big( h(x) - \lambda_i(x) \Big)^2 \\
     & = \frac{1}{a_p + a_n} \Big( a_p (h(x) - 1)^2 + a_n h(x)^2 \Big) \\
     & = \frac{1}{a_p + a_n} \Big( a_p h(x)^2 - 2a_p h(x) + a_p + a_n \cdot h(x)^2 \Big) \\
     & = h(x)^2 - 2 \frac{a_p}{a_p + a_n} h(x) + \frac{a_p}{a_p + a_n} \\
     & = h(x)^2 - 2 \frac{a_p}{a_p + a_n} h(x) + \big( \frac{a_p}{a_p + a_n} \big)^2 + \\
     & \quad \quad \quad \frac{a_p}{a_p + a_n} - \big( \frac{a_p}{a_p + a_n} \big)^2\\
     & = \big( h(x) - \frac{a_p}{a_p + a_n} \big)^2 + \frac{a_p}{a_p + a_n} - \big( \frac{a_p}{a_p + a_n} \big)^2\\
     & = \ell_{\text{square}} \Big(h(x), \frac{a_p}{a_p + a_n} \Big) + \frac{a_p}{a_p + a_n} - \big( \frac{a_p}{a_p + a_n} \big)^2.
\end{align*}
\normalsize
\end{proof}

Again, we have that the LoL objective is similar to using the soft majority vote as a pseudolabel, under the square loss. There is an additional term of $\big( \frac{a_p}{a_p + a_n} - \big( \frac{a_p}{a_p + a_n} \big)^2\big)$, but we note that this term is a constant with respect to $h$ (our model), and thus does not have any influence on optimization. Therefore, a simple aggregation of these losses recovers the same objective as using an intermediate pseudolabel. However, we can incorporate \textbf{more} information than what pseudolabels naturally allow by incorporating the gradient of the heuristics into our loss functions directly.

\section{Additional Experimental Details} \label{exp_details}
We perform hyperparameter optimization of all methods with Optuna \citep{optuna}, selecting the best set of parameters on the validation set. We optimize over the following parameters:
\begin{itemize}
    \item learning rate: $\{0.1, 0.01, 0.001, 0.0001\}$
    \item weight decay: $\{0, 0.01, 0.001\}$
    \item gradient threshold ($c$): $[0, 5]$
    \item gradient weight ($\alpha$): $\{ 0.1, 0.01, 0.001, 0.0001, 0.00001 \}$
\end{itemize}
In all of our experiments, we use a fixed architecture of a 3-layer MLP with hidden dimensions [64, 16] and a ReLU activation function. We use ReLU activations and Dropout layers \citep{dropout} between our hidden layers, with probability $p = 0.2$. We use the Adam optimizer \citep{adam} and train all models for 30 epochs with a batch size of 128. We note that the last two hyperparameters (the gradient threshold and gradient weighting) are only used by LoL.

\section{Weighted Combination of Losses} \label{weightings}

We provide experiments to demonstrate that our method can be further improved by weighting our combination of losses. Our approach in Equation \ref{gradloss} still takes a naive, unweighted combination of these losses. There are many potential ways to combine these losses, such as through a notion of weak labelers' importance. As we provide no weighting in our combination, weak labelers are implicitly weighted differently based on their abstention rates. For example, if one weak labeler abstains on 50\% of the data while the other weak labeler does not abstain, then the first weak labeler only contributes 25\% to the overall loss. Therefore, a natural way to weight these losses is to have them contribute equal weight to the overall objective (i.e, re-weighting by their coverage). We denote $p_i$ as the number of times of $\lambda_i$ votes, or $p_i = \sum_{j=1}^n \mathbbm{1}\{\lambda_i(x_j) \neq \emptyset \}$. This produces the following objective
\begin{equation}\label{autoloss-eq}
    \hat{h}_c = \argmin_h \: \sum_{j=1}^n \Bigg( \frac{1}{m(x_j)} \cdot \sum_{i=1}^m \frac{\ell_i(x_j, h)}{p_i} \Bigg).
\end{equation}

The standard approach in crowdsourcing and weakly supervised learning is to combine weak labelers by estimates of their accuracies. 
To remain in an unlabeled setting, we follow the same procedure as presented in Snorkel MeTaL \citep{multi_task_weak_supervision} to produce accuracy estimates. 
We use their default graphical model, which is described in greater detail in Appendix \ref{snorkel_weights}. Given a set of accuracies $w = \{w_1, ..., w_m\}$, we present the following optimization problem to find an optimal accuracy-weighted classifier $\hat{h}_a$:
\begin{equation}\label{swloss}
    \hat{h}_{a} = \argmin_h  \sum_{i=1}^m \Bigg( \frac{1}{m(x_j)} \sum_{j=1}^n  \left(\frac{w_i}{z}\right) \ell_i(x_j, h) \Bigg),
\end{equation}
where $z = \sum_{l=1}^k w_l$. This method and the coverage-based weighting provided in the main text are concrete examples of simple and intuitive aggregations of our loss functions. We provide empirical results to analyze the impact of different weighted combinations of our losses. We denote the solution to optimizing the weighted combinations by coverage (Equation \ref{autoloss-eq}) and by accuracy estimates (Equation \ref{swloss}) as \textbf{LoL-c} and \textbf{LoL-a} respectively. We add Snorkel (MeTaL) as a comparison, as we generate the accuracy estimates through Snorkel's graphical model. We use the same experimental setup as in generating Table \ref{tab:results}.

\renewcommand{\arraystretch}{1.4}
\begin{table}[h]
\centering
\caption{Results for different weighting schemes for LoL when trained on \textit{all} of the unlabeled WRENCH training data, averaged over 5 random seeds. Accuracies are reported as mean $\pm$ standard deviation. In the first row, we denote our methods in bold (LoL, LoL-c, LoL-a). We bold the best-performing method on each dataset.}
\vspace{2mm}
\setlength{\tabcolsep}{5pt}
\resizebox{0.98\columnwidth}{!}{
\begin{tabular}{l | c | c c c}
\toprule
 & Snorkel & \textbf{LoL} & \textbf{LoL-c} & \textbf{LoL-a} \\ \midrule
agnews      & 82.7 $\pm$ 0.9 & 83.4 $\pm$ 0.1 & \textbf{83.9 $\pm$ 0.2} & 83.6 $\pm$ 0.1 \\
chemprot    & 51.0 $\pm$ 0.4 & \textbf{52.9 $\pm$ 0.3} & 51.2 $\pm$ 0.3 & 52.4 $\pm$ 0.3 \\
IMDB        & \textbf{82.1 $\pm$ 0.5} & 81.8 $\pm$ 0.3 & 75.0 $\pm$ 0.3 & 81.6 $\pm$ 0.6 \\
Yelp        & \textbf{77.7 $\pm$ 0.5} & 75.9 $\pm$ 0.7 & 76.5 $\pm$ 0.6 & 76.0 $\pm$ 0.6 \\
YouTube     & 91.0 $\pm$ 0.3 & 94.2 $\pm$ 0.7 & \textbf{95.0 $\pm$ 0.6} & 93.8 $\pm$ 0.7  \\ \bottomrule
\end{tabular}
}
\label{tab:weighting_results}
\end{table}

A surprising result is that the accuracy-based weighting (in an unsupervised fashion via Snorkel) does not seem to benefit our approach on these text classification tasks, likely due to inaccuracies in estimating weak labeler accuracies. However, results on our image classification experiments differ as Snorkel is able to accurately estimate weak labeler accuracies. As a result, both coverage-based weighting and accuracy-based weightings are simple and natural extensions of our method that can lead to potential improves upon the standard LoL aggregation. 

\section{Accuracy Estimatition} \label{snorkel_weights}

Here, we briefly present the necessary ideas and notation from  Snorkel MeTaL \citep{multi_task_weak_supervision}, which is used to produce accuracy estimates in an unsupervised fashion. We then use these accuracies as weights to combine our losses derived from weak labelers. We create a graphical model $P_\mu(Y, \lambda)$, which corresponds to a joint distribution over the true labels and our weak labelers $\lambda$. We remark that $Y$ is \textit{unobserved} during training, as we are in an unsupervised setting. We also observe some user-specified graph $G = \{Y, \lambda_1, ..., \lambda_m \}$. The presence of an edge in this graph denotes a dependency among two vertices, although the default setting (and what we use in our experiments) is a completely disconnected graph; i.e., making the class conditional independence assumption. Learning these edges without prior knowledge is a completely separate problem setting \citep{learning_structure}.

We denote $\mathcal{C}$ as the set of maximal cliques in the graph $G$. Then for some $C \in \mathcal{C}$, we define $\mu = E[ \psi(C)]$, where $\psi(C) \in \{0, 1 \}^{\prod_{i \in C}(|\mathcal{Y}| - 1)}$ represents the vector of indicator variables for all combinations of labels except one for each node in the clique $C$. Then, $\mu$ corresponds to a vector of sufficient statistics for the graphical model. To solve for $\mu$ and consequently $P_\mu(Y, \lambda)$, we can use their covariance matrix completion objective; we defer interested readers to their original work \citep{multi_task_weak_supervision} for algorithmic details and theoretical analysis. 

Then, we can use these statistics $\mu$ to compute the weak labeler accuracy $w_i$ of weak labeler $\lambda_i$ as 
\begin{equation*}
    w_i = \frac{\sum_{y \in \mathcal{Y}} P_\mu(\lambda_i = y, Y = y) \cdot P_\mu(Y = y)}{P_\mu(\lambda_i \neq \emptyset)}
\end{equation*}

which simply sums up the probabilities where the weak labeler matches the true unobserved label, divided by its coverage. We remark that this graphical model makes strong assumptions, namely the class-conditional independence assumption and equal class balance (that $P_\mu(Y = y) = \frac{1}{|\mathcal{Y}|}, \forall y \in \mathcal{Y}$). Therefore, these weightings (and consequently LoL-a and Snorkel) are potentially inaccurate when these assumptions are violated. 

\section{Comparing Gradient Penalties} \label{grad_ablation}

We provide an ablation study to compare different potential choices of gradient penalties in our loss generation process. We denote a square penalty (the method used in the main body of the paper) as \textbf{LoL}. We can also consider a linear penalty and an exponential penalty, which we refer to as \textbf{LoL-linear} and \textbf{LoL-exp} respectively. Formally, these are given by
\small
\begin{align*}
    \ell^{\text{linear}}_i(x, h) = & \:   \mathbbm{1} \{\lambda _i(x) \neq \emptyset \} \cdot \Big( \ell(h(x), \lambda_i(x)) + \\
    & \quad \alpha \cdot || \Big[ c  \cdot \nabla_{\phi} \tilde{\lambda}_i(\phi)_\mathcal{Y} - \nabla_{\phi}  \tilde{h}(\phi) \Big]^+ ||_{1} \Big) \\
    \ell^{\text{exp}}_i(x, h) = & \:   \mathbbm{1} \{\lambda _i(x) \neq \emptyset \} \cdot \Big( \ell(h(x), \lambda_i(x)) + \\
    & \quad \alpha \cdot \sum_{j=1}^d \exp \big( \Big[ c  \cdot \nabla_{\phi} \tilde{\lambda}_i(\phi) _\mathcal{Y} - \nabla_{\phi}  \tilde{h}(\phi) \Big]^+_{j} \big) \Big),
\end{align*}
\normalsize
where our data $x$ is $d$-dimensional. The main difference between these approaches is the amount of gradient penalty placed on the model; the exponential penalty results in a larger loss when the model has gradients that are more different than that of the heuristics, in comparison to both the square and linear penalties. We run these methods in the same experimental setting on the WRENCH benchmark and report the results in Table \ref{tab:pen_results}.

\begin{figure*}[t]
    \centering
    \includegraphics[width=0.64\columnwidth]{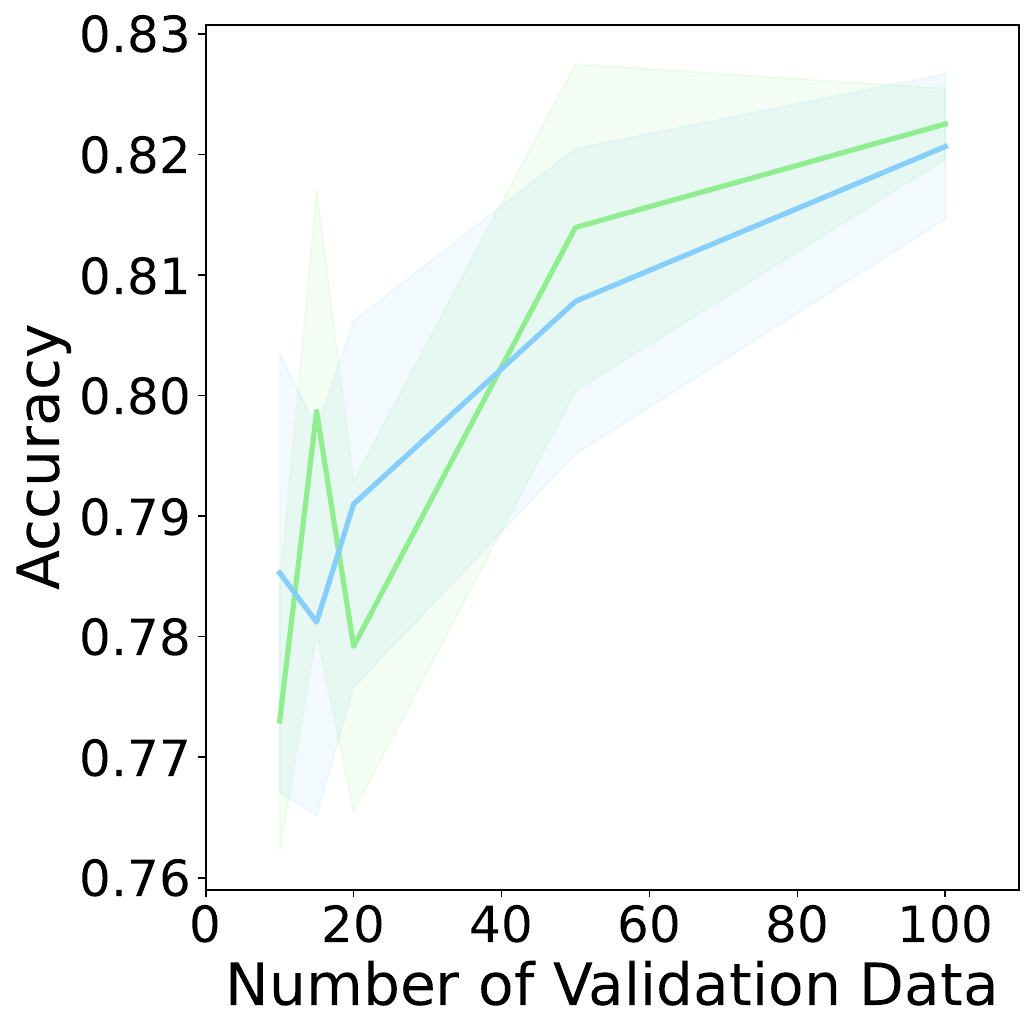}
    \includegraphics[width=0.64\columnwidth]{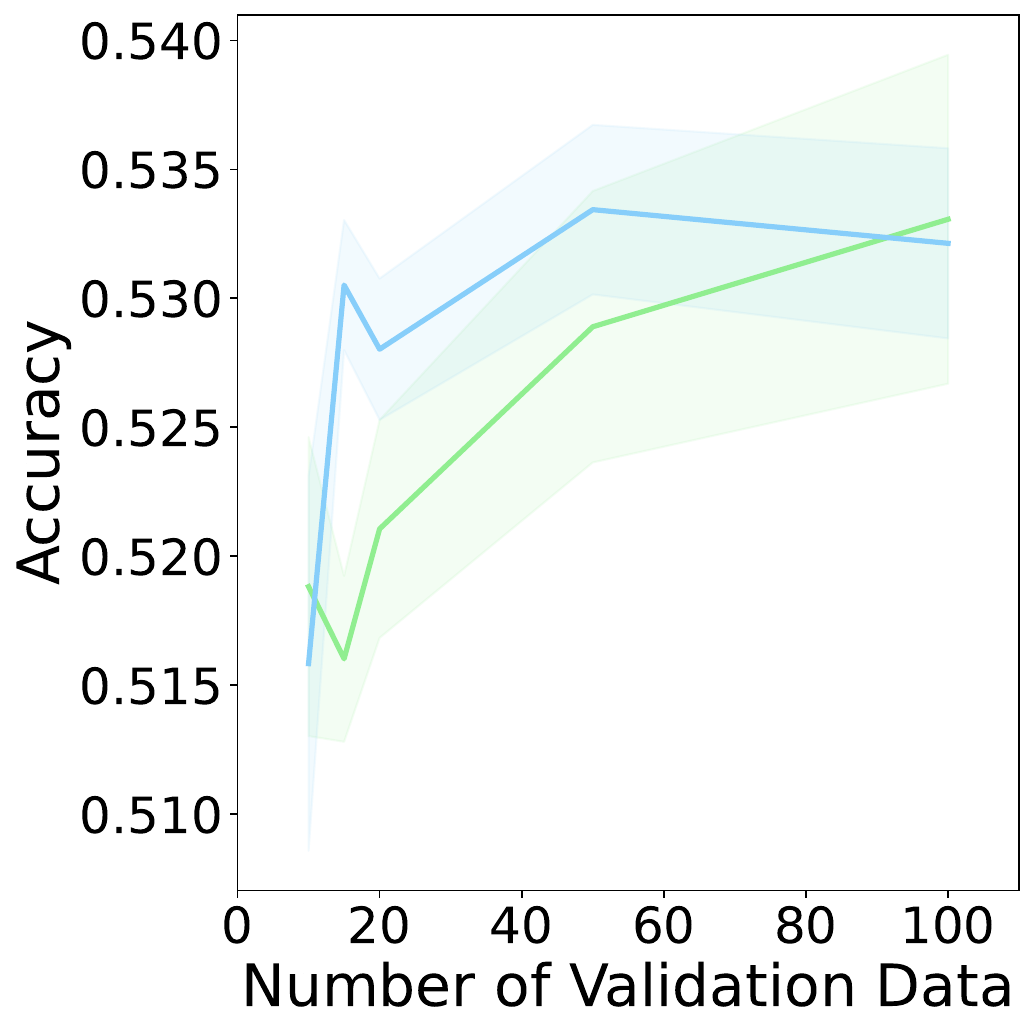} 
    \includegraphics[width=0.64\columnwidth]{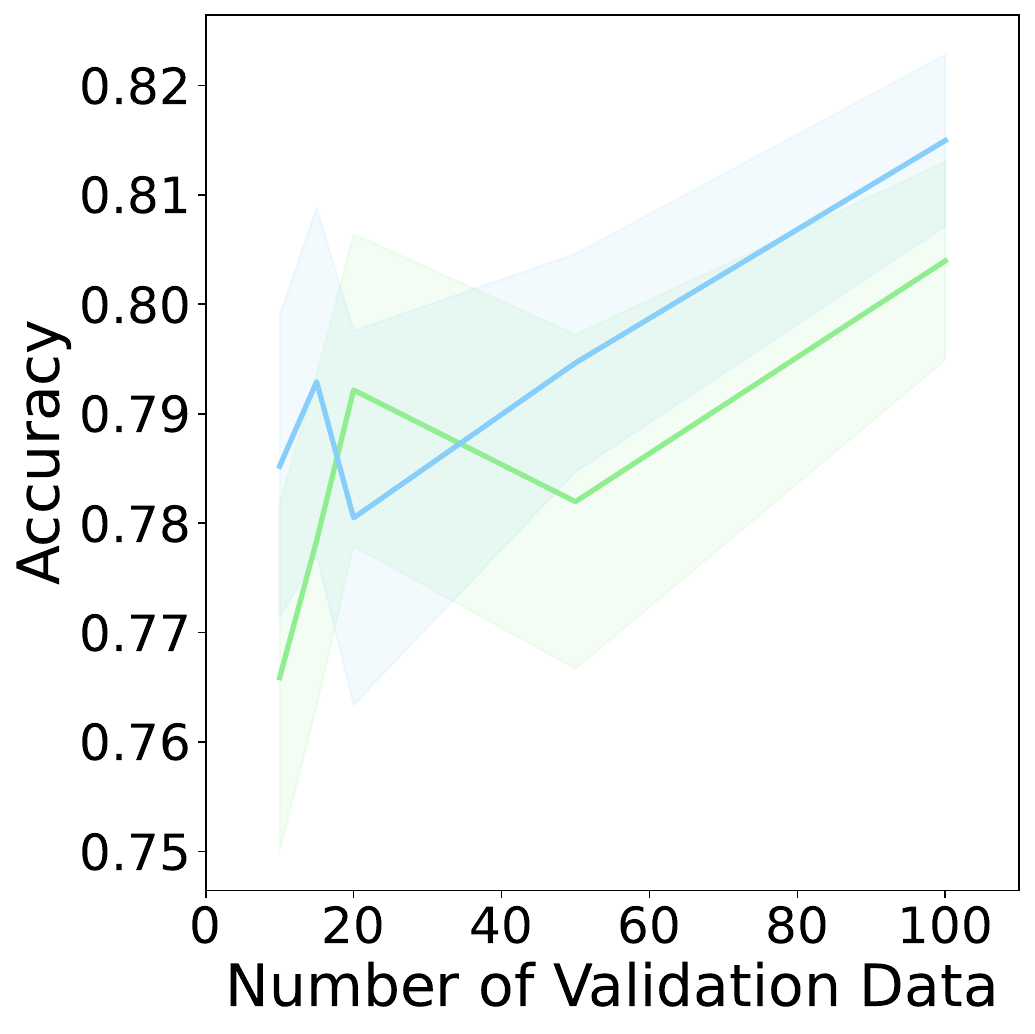}
    \includegraphics[width=0.64\columnwidth]{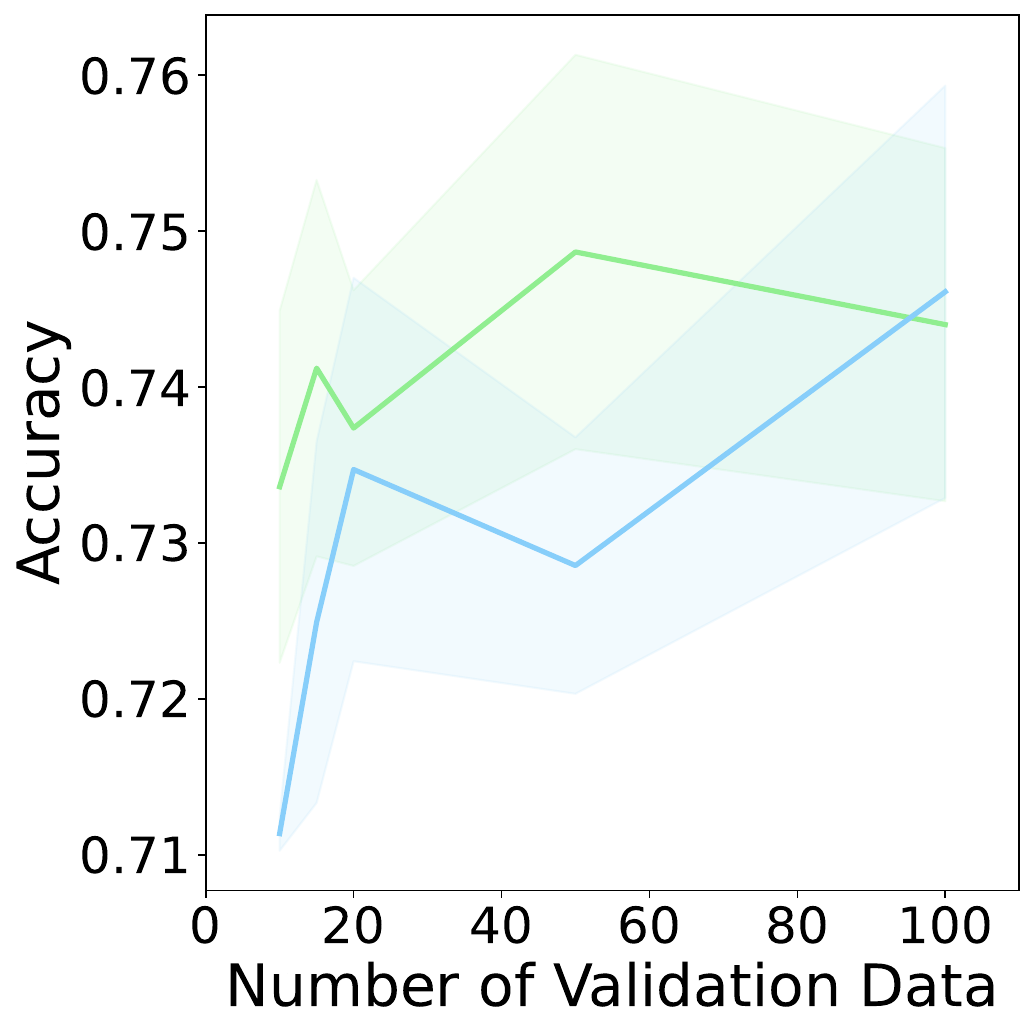}
    \includegraphics[width=0.64\columnwidth]{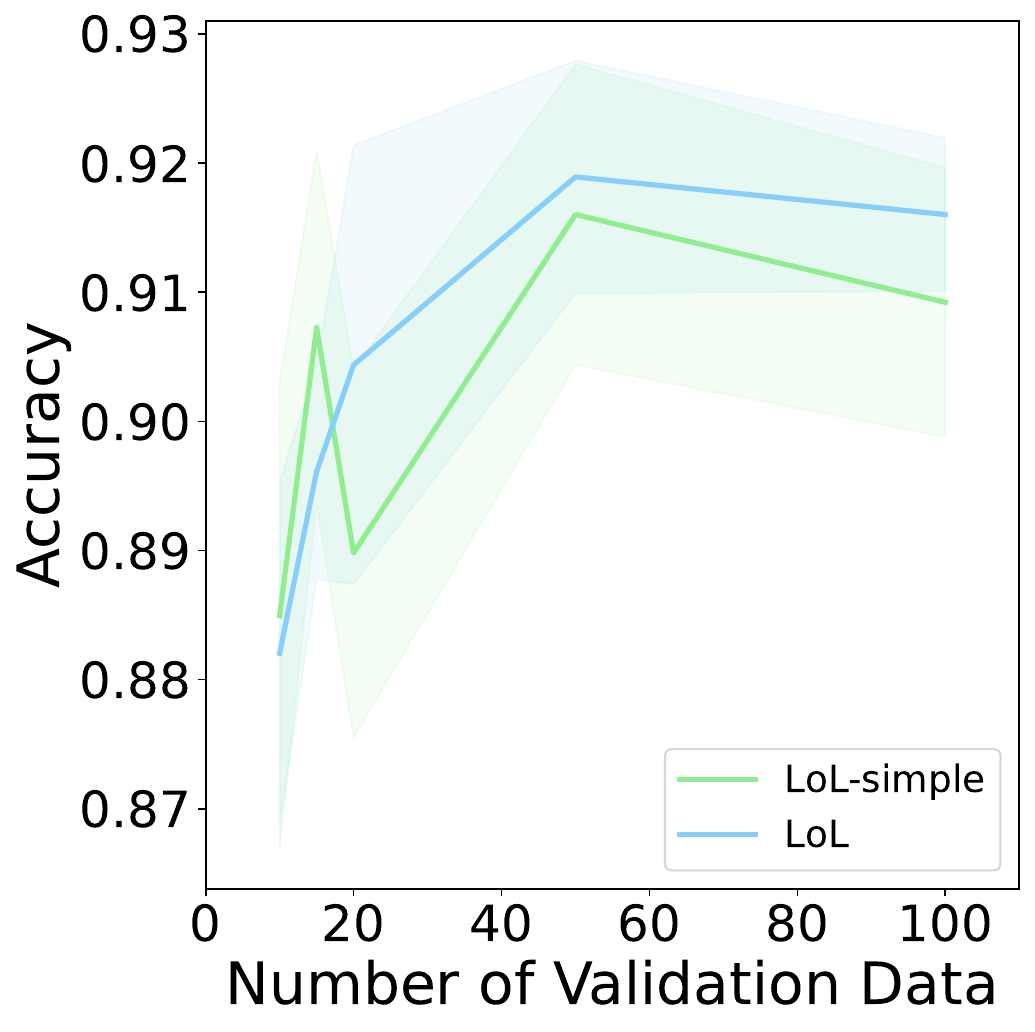}
    \caption{Results (from left to right and top to bottom: agnews, chemprot, IMDB, Yelp, Youtube) as we vary the size of the validation set for LoL and LoL-simple. We compute accuracies across 5 random seeds to compute error bars ($ \pm $ standard deviation) with validation datasets of sizes $n=10, 15, 20, 50, 100$.}
    \label{fig:vary_lab_grad}
\end{figure*}

\renewcommand{\arraystretch}{1.4}
\begin{table}[h]
\centering
\caption{Results on 5 text classification datasets from the WRENCH benchmark, averaged over 5 random seeds. Accuracies are reported as mean $\pm$ standard deviation.}
\vspace{2mm}
\setlength{\tabcolsep}{5pt}
\resizebox{0.8\columnwidth}{!}{
\begin{tabular}{l | c c c}
\toprule
 & \textbf{LoL} & \textbf{LoL-linear} & \textbf{LoL-exp} \\ \midrule
agnews    & \textbf{83.4 $\pm$ 0.1} & 83.3 $\pm$ 0.2 & 83.3 $\pm$ 0.1  \\
chemprot  & 52.9 $\pm$ 0.3 & 53.7 $\pm$ 0.5 & \textbf{54.2 $\pm$ 0.8} \\
IMDB      & \textbf{81.8 $\pm$ 0.3} & 81.4 $\pm$ 0.7 & 81.6 $\pm$ 0.9 \\
Yelp      & 75.9 $\pm$ 0.7 & 75.4 $\pm$ 0.8 & \textbf{77.1 $\pm$ 2.4} \\
YouTube   & \textbf{94.2 $\pm$ 0.7} & 94.1 $\pm$ 0.8 & 93.7 $\pm$ 1.2  \\ \bottomrule
\end{tabular}
}
\label{tab:pen_results}
\end{table}

Our results do not show a clear best penalty; however, it does illuminate that other gradient penalties can improve our LoL method. We observe that the exponential penalty noticeably improves over the square penalty on both the chemprot and Yelp datasets. We also note that this penalty seems to lead to a solution with higher variance. The linear penalty, however, seems to perform worse than the square penalty on all datasets except chemprot. This result shows that there can be further improvements to the LoL method, and future work can look into different gradient matching objectives.

\section{Impact of Using Gradient Information with Limited Unlabeled Data} \label{lol_vs_lol-simple_ul}

We again provide experiments to compare LoL and LoL-simple as we vary the amount of validation data used since LoL has a larger number of hyperparameters. We consider the same experimental setting as used for Figure \ref{fig:vary_lab_grad}. Although the LoL method has a larger number of hyperparameters, we observe similar or better performance of LoL when compared to the LoL-simple method in the limited labeled data regime. In fact, the LoL method starts to outperform the method on most datasets with only 100 labeled data points for validation.


\section{Comparison Against Methods with Additional Information}

As mentioned previously in our related work section, many recent methods in weak supervision assume the presence of additional information. One common form of this additional information is a prior notion of the accuracy of the weak labelers, which our method and other baselines in the main text do not require. We provide experimental results as we compare against the following method.

\textbf{Constrained Label Learning \citep{Arachie2021ConstrainedLF}}: We compare against an approach that produces a set of pseudolabels from a feasible space of possible labelings, which is constrained by weak labelers accuracies. Their code is released for binary classification tasks only, so we report their performance on our binary classification datasets. We compare against this method that uses the \textit{true weak labeler accuracies} (i.e., a method requiring more information) and refer to it as \textbf{CLL-True}. 
To have a more fair comparison, we also compare a version of their method that uses a constant error rate, which we refer to as \textbf{CLL-Constant}. We cannot use their agreement-based method as these datasets have weak labelers that abstain. As in their experimental section, we use an constant error rate of $\epsilon = 0.01$ on all of the (text) datasets. We again compare LoL to these methods on the WRENCH benchmark, both with the full unlabeled training data (Table \ref{tab:cll_all}) and with only 100 unlabeled training points (Table \ref{tab:cll_ul}).

\renewcommand{\arraystretch}{1.4}
\begin{table}[h]
\centering
\caption{Results when comparing LoL to methods that use constraint-based methods on binary classification tasks, averaged over 5 random seeds. Accuracies are reported as mean $\pm$ standard deviation. We bold the best performing method. We note that CLL-True uses \textbf{more information} that the other methods.}
\vspace{2mm}
\setlength{\tabcolsep}{5pt}
\resizebox{0.9\columnwidth}{!}{
\begin{tabular}{l | c c | c}
\toprule
 & CLL-Constant & CLL-True & \textbf{LoL} \\ \midrule
IMDB     & 80.9 $\pm$ 0.4 & 80.3 $\pm$ 0.3 &  \textbf{81.8 $\pm$ 0.3}  \\
Yelp     & \textbf{79.1 $\pm$ 1.3} & 76.9 $\pm$ 0.7 & 75.9 $\pm$ 0.7 \\
YouTube  & 92.5 $\pm$ 0.4 & \textbf{94.3 $\pm$ 0.8} &  94.2 $\pm$ 0.7 \\ \bottomrule
\end{tabular}
}
\label{tab:cll_all}
\end{table}

\renewcommand{\arraystretch}{1.4}
\begin{table}[h]
\centering
\caption{Results when only training on \textit{100 unlabeled data}, averaged over 5 random seeds. Accuracies are reported as mean $\pm$ standard deviation. Again, we bold the best-performing method.}
\vspace{2mm}
\setlength{\tabcolsep}{5pt}
\resizebox{0.9\columnwidth}{!}{
\begin{tabular}{l | c c | c }
\toprule
 & CLL-Constant & CLL-True & \textbf{LoL} \\ \midrule
IMDB      & 67.5 $\pm$ 0.3 & 68.0 $\pm$ 0.5 & \textbf{69.6 $\pm$ 0.4}\\
Yelp      & 66.3 $\pm$ 2.2 & 70.0 $\pm$ 1.0 & \textbf{71.2 $\pm$ 2.4} \\
YouTube   & 90.3 $\pm$ 1.6 & 91.9 $\pm$ 0.9 & \textbf{92.0 $\pm$ 0.9} \\ \bottomrule
\end{tabular}
}
\label{tab:cll_ul}
\end{table}

When using the full training data, we remark that our method performs better than CLL-Constant in 2 of the 3 datasets and CLL-True on 1 of the datasets. In addition, our LoL method is comparable to CLL-True on the YouTube dataset.
In the limited unlabeled data regime (with only 100 unlabeled points), we observe that our method outperforms both methods (even with the true weak labeler accuracies), similarly to how it outperformed all other baselines. These experiments further demonstrate the applicability of LoL, as it performs comparably to methods that use additional information (CLL-True), especially outperforming all other methods when there is limited training data.

\section{Additional Details}\label{compute_res}

We use cluster compute resources to produce our empirical results. We use a single GPU (NVIDIA GeForce RTX 2080Ti) to run our LoL method and each of the baselines.

We use the WRENCH benchmark \citep{wrench} for all of our data. This benchmark has a Apache-2.0 license.


\end{document}